\documentclass[letterpaper, 10 pt, journal, twoside]{IEEEtran}  

\IEEEoverridecommandlockouts                              

\usepackage{graphicx} 
\usepackage{subcaption}
\usepackage{epsfig} 
\usepackage{amsmath} 
\usepackage{amssymb}  
\usepackage{float}
\usepackage{multirow}
\usepackage{siunitx}
\usepackage{makecell}
\usepackage{hyperref}
\usepackage[dvipsnames]{xcolor}
\usepackage{pifont}
\newcommand{\cmark}{\ding{51}}%
\newcommand{\xmark}{\ding{55}}%
\begin{document}

\title{Learning a State Representation and Navigation in Cluttered and Dynamic Environments}

\author{David Hoeller$^{1,2}$, Lorenz Wellhausen$^{1}$, Farbod Farshidian$^{1}$, Marco Hutter$^{1}$
\thanks{Manuscript received: October, 15, 2020; Revised January, 18, 2021; Accepted February, 22, 2021.}
\thanks{This paper was recommended for publication by Editor Nancy Amato upon evaluation of the Associate Editor and Reviewers' comments.}
\thanks{This work was supported by NVIDIA, the Swiss National Science Foundation (SNSF) through project 166232, 188596, the National Centre of Competence in Research Robotics (NCCR Robotics), and the European Union's Horizon 2020 research and innovation program under grant agreement No.780883. Moreover, this work has been conducted as part of ANYmal Research, a community to advance legged robotics.}%
\thanks{$^{1}$ All authors are with the Robotic Systems Lab, ETH Z\"u{}rich, Switzerland {\tt\footnotesize dhoeller@ethz.ch}}%
\thanks{$^{2}$ D. Hoeller is also with NVIDIA}%
\thanks{Digital Object Identifier (DOI): see top of this page.}
}

\markboth{IEEE Robotics and Automation Letters. Preprint Version. Accepted February, 2021} {Hoeller \MakeLowercase{\textit{et al.}}: Learning a State Representation and Navigation in Cluttered and Dynamic Environments}

\maketitle

\begin{abstract}

In this work, we present a learning-based pipeline to realise local navigation with a quadrupedal robot in cluttered environments with static and dynamic obstacles. 
Given high-level navigation commands, the robot is able to safely locomote to a target location based on frames from a depth camera without any explicit mapping of the environment.
First, the sequence of images and the current trajectory of the camera are fused to form a model of the world using state representation learning. The output of this lightweight module is then directly fed into a target-reaching and obstacle-avoiding policy trained with reinforcement learning. 
We show that decoupling the pipeline into these components results in a sample efficient policy learning stage that can be fully trained in simulation in just a dozen minutes. The key part is the state representation, which is trained to not only estimate the hidden state of the world in an unsupervised fashion, but also helps bridging the reality gap, enabling successful sim-to-real transfer. 
In our experiments with the quadrupedal robot ANYmal in simulation and in reality, we show that our system can handle noisy depth images, avoid dynamic obstacles unseen during training, and is endowed with local spatial awareness.

\end{abstract}


\section{INTRODUCTION}
\IEEEPARstart{N}{avigating} in dynamic environments with an autonomous robot is a complicated task. Dynamic objects have to be singled out from raw sensor data, their motion predicted into the future, and this information carried over to the controller \cite{4621214}. The latter needs to compute a collision-free trajectory fast enough to react to dynamic obstacles and be robust to noise in the measurements. 
This problem requires the efficient combination of spatio-temporal filtering and trajectory optimization. 
However, sensors such as LiDARs and depth cameras provide a large amount of mostly redundant data and processing these data fast enough to avoid moving objects usually requires a tailored hardware-software solution.

The typical model-based pipeline either assumes a static environment and relies on the controller reacting fast enough to oncoming obstacles \cite{dwa, 7743420}, 
or have a reactive controller that directly integrates velocity measurements from time of flight sensors into the control formulation \cite{44033}.
Another approach is to fit a velocity model on dynamic objects in the scene and use that information in the cost map of a trajectory optimizer \cite{4621214, obstacle}. The challenge with such methods is to correctly estimate the evolution environment representation (such as occupancy grids or signed distance fields) in the optimizer. Therefore, such pipelines quickly become complicated, resort to heuristics, and suffer from time delays coming from signal processing. 

On the other hand, deep learning-based approaches can extract useful information from large unlabeled datasets, removing the need for feature engineering or heuristics. 
Moreover, deep Reinforcement Learning (RL) has achieved remarkable progress in recent years in the field of continuous control \cite{lillicrap2016RL,hwangbo2019RL}. For that reason, new research has emerged to solve the navigation problem using learning \cite{SadeghiL17, pfeiffer2017perception, tai2018social, 8461113}. However, few approaches train the policy in simulation with synthetically generated images \cite{SadeghiL17, kaufmann2020RSS}, mainly because simulators suffer from the reality gap \cite{Tobin2017DomainRF}. Nevertheless, simulation engines are a promising tool to tackle this problem because they can generate a great variety of training scenarios very quickly. 

\begin{figure}[t]
    \centering
    \includegraphics[width=1.0\columnwidth]{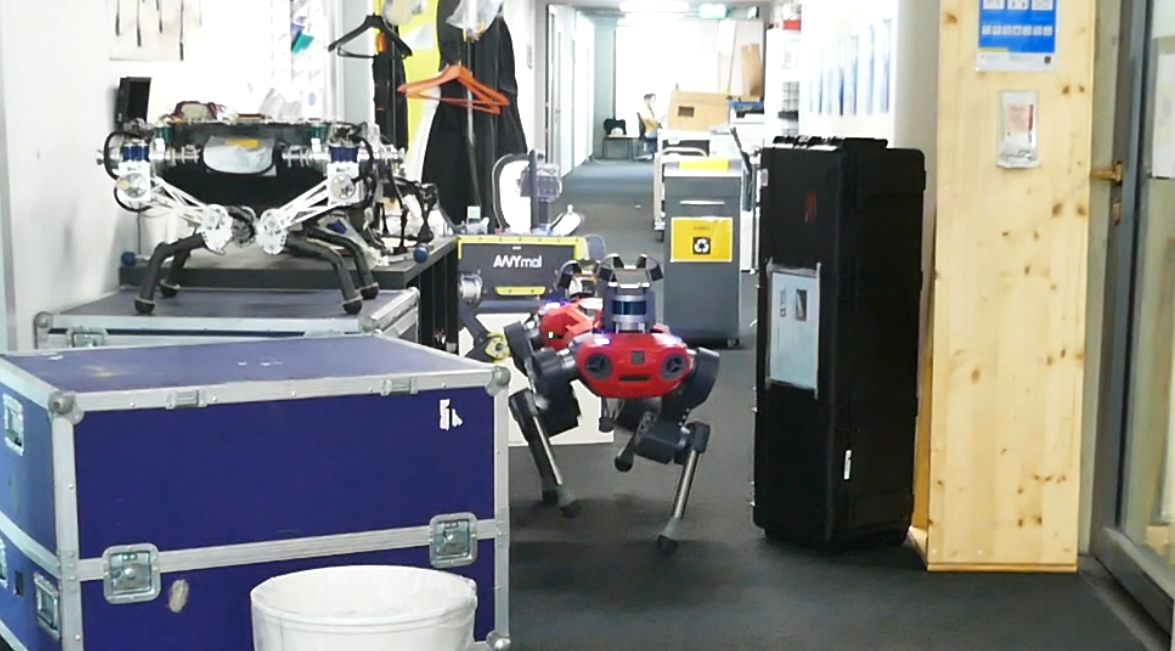}
    \caption{The quadrupedal robot ANYmal navigating in a cluttered environment with tight spaces and dynamic obstacles. }
    \label{fig:intro}
    \vspace*{-6mm}
\end{figure}

\begin{figure*}[t]
\vspace*{5mm}
\centering
\includegraphics[width=0.9\textwidth]{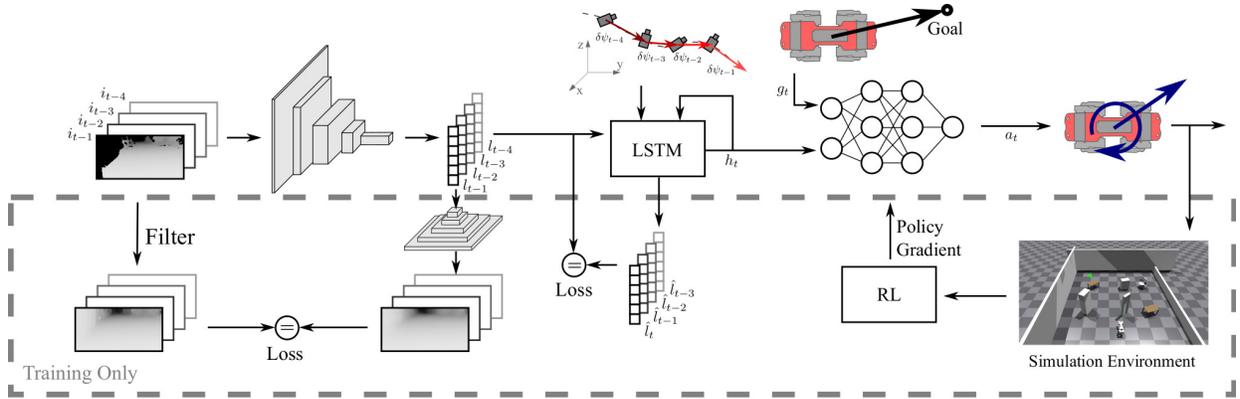}
\caption{Description of our approach. A sequence of images is fed through an encoder to produce latent vectors. These vectors are fed through an LSTM network along with the camera's trajectory to learn a representation of the world. That representation is used by the policy trained in simulation to avoid obstacles.}
\label{fig:pipeline}
\vspace*{-6mm}
\end{figure*}

\subsection{Contribution}
This paper presents a learning-based method to navigate locally in dynamic environments with a mobile robot. At every time-step, the robot receives the current depth frame from the front-facing camera, the relative coordinates of a nearby goal to reach, and decides on a velocity command. The robot's goal is to successfully navigate towards the goal, while avoiding potential hazards along the way. 

To achieve this, we develop a standalone module that can predict the scene's temporal evolution from a stream of depth images and the camera's trajectory. This module estimates the partially observable state of the world without having to detect dynamic objects and explicitly predict their motion, and does not rely on heuristics. This module can then be used in a model-based or a learning-based control setting. Here, we directly use that module's latent space to train an obstacle avoiding policy on a mobile robot entirely in simulation using RL. We deploy the policy to the quadrupedal robot ANYmal~\cite{anymal} sim-to-real and show that it can navigate in a cluttered scene and avoid people moving across its path in a very reactive manner.

The approach depicted in Fig.~\ref{fig:pipeline} can complement SLAM-based techniques and handle dynamic obstacles, which are typically not captured by such methods. 
Indeed, using a map built with SLAM, a global path can be computed and these waypoints given to our pipeline to locally adapt the path to avoid dynamic obstacles and new obstacles that were not there during the global path planning stage.




To summarize, the main contributions of this work are the development of a task and robot agnostic state representation module that can operate in simulation and reality, generalizes to unseen objects, uses no heuristics, and the design of a sample efficient sim-to-real RL pipeline using that representation to navigate in cluttered and dynamic scenes. 



\subsection{Related Work}
\label{sec:related}
Already in 2005, the authors of \cite{SFnav} proposed a system that uses a monocular camera to estimate the distance to obstacles geometrically and use that information to train a policy in simulation. Other works feed RGB images or 2D LiDAR data through a neural network to produce a control command for navigation by learning from expert trajectories \cite{Kaufmann2018DeepDR, DroNet, forestIdsia, pfeiffer2017perception}, via reinforcement learning \cite{xie2017towards, Everett2018MotionPA}, or fully self-supervised \cite{Badgr}. In this work, we do not assume the presence of an expert and use depth images.
RGB-based methods such as \cite{visNavOC, mueller2018driving} do not explicitly consider dynamic environments in their formulation, resulting in purely reactive policies. More specifically, the authors in \cite{SadeghiL17, Sadeghi19} use a network to estimate whether there is an obstacle 1 meter ahead of the robot from monocular RGB images. As a result, an obstacle moving fast towards the robot from a distance will only be detected when it is close to the robot, and the robot might not have time to react. 
Also, using depth images instead of RGB in a simulation-based learning setting results in a smaller reality gap and reduces the complexity of the data collection pipeline. In comparison, the authors in \cite{SadeghiL17, Sadeghi19} use RGB images and had to design 24 worlds in game engines manually, use 200 textures, and 21 different furniture items. Here, we simply use four furniture items and place them randomly across the room. LiDAR-based methods such as \cite{AutoRL, pedestrianNav} do not use 3D LiDARs. Therefore, smaller obstacles on the ground cannot be detected. 
Contrary to most of the works mentioned above, we explicitly consider dynamic obstacles in our formulation. We focus on estimating the scene's temporal evolution in a standalone module and use this to navigate without collisions. 








A state representation is a rich signal that has shown to be useful for control purposes \cite{LESORT2018379}. In the World Models approach \cite{NIPS2018_7512}, the representation is used to play games from pixels. They use a simple control formulation to map the representation to actions and use an evolutionary algorithm to train the policy. Other works learn a latent representation, which is then used with model-based control \cite{E2C, hafner2019planet} or RL \cite{lee2019slac}. A suitable representation for control can also be generated from neural rendering \cite{Eslami2018NeuralSR, NIPS2019_9331}, where a new viewpoint of a scene can be generated from other viewpoints. However, none of these works have shown results on a real robot.

\section{Learning a representation of the world}
\label{sec:state_representation}
Learning a policy from pixels directly is a challenging task, and it has been shown empirically that feeding a state signal instead results in better sample efficiency \cite{laskin_srinivas2020curl}.

In this Section, we describe how to estimate the state of the world from pixel measurements. The resulting low-dimensional but rich signal can be used in an RL setting to efficiently learn the navigation task.

More precisely, since we do not know the world's configuration and the motion of the obstacles a priori, we are dealing with a partially observable problem. The agent obtains noisy measurements in the form of depth images $d_t$ and the camera's trajectory $\delta \psi_{t}$. These narrow snapshots of the world must be fused together to estimate the world's underlying hidden state.
\subsection{Encoding depth images}
\label{sec:VAE}
\begin{figure*}[t!]
    \vspace*{5mm}
    \centering
    \setlength\lineskip{3pt}
    \hspace*{\fill}
    \begin{subfigure}[t]{0.24\textwidth}
        \centering
        \includegraphics[width=\textwidth]{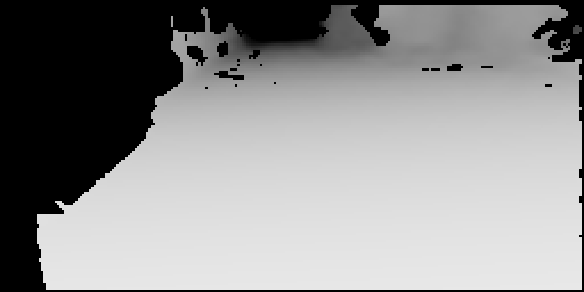}
    \end{subfigure} \hfill
    \begin{subfigure}[t]{0.24\textwidth}
        \centering
        \includegraphics[width=\textwidth]{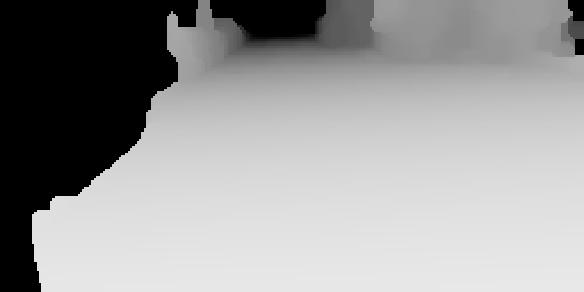}
    \end{subfigure} \hfill
    \begin{subfigure}[t]{0.24\textwidth}
        \centering
        \includegraphics[width=\textwidth]{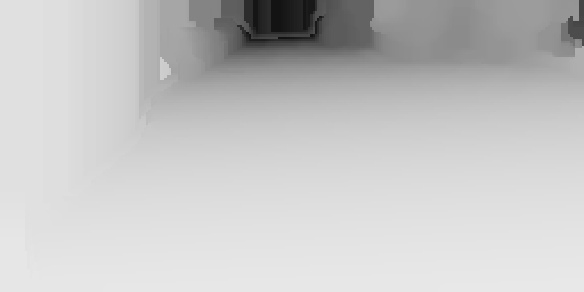}
    \end{subfigure} \hfill
    \begin{subfigure}[t]{0.24\textwidth}
        \centering
        \includegraphics[width=\textwidth]{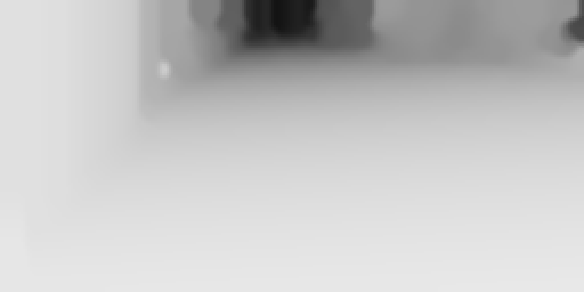}
    \end{subfigure}
    \hspace*{\fill}
    
    \hspace*{\fill}
    \begin{subfigure}[t]{0.24\textwidth}
        \centering
        \includegraphics[width=\textwidth]{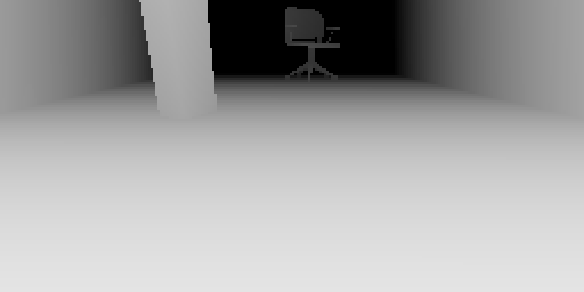}
        \caption{Input}
    \end{subfigure} \hfill
    \begin{subfigure}[t]{0.24\textwidth}
        \centering
        \includegraphics[width=\textwidth]{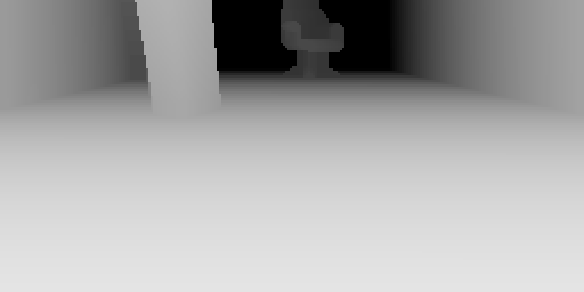}
        \caption{\centering Dilation and \newline hole closing}
    \end{subfigure} \hfill
    \begin{subfigure}[t]{0.24\textwidth}
        \centering
        \includegraphics[width=\textwidth]{images/morphology_sim.png}
        \caption{Large dilation}
    \end{subfigure} \hfill
    \begin{subfigure}[t]{0.24\textwidth}
        \centering
        \includegraphics[width=\textwidth]{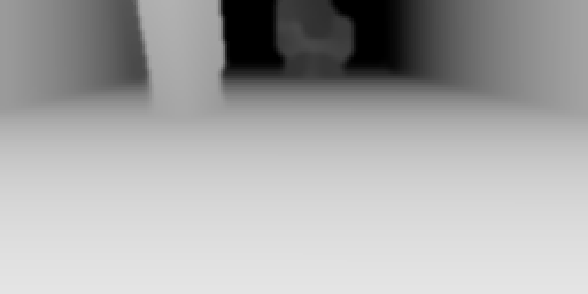}
        \caption{Bilateral filtering}
    \end{subfigure}
    \hspace*{\fill}
    \caption{Different filtering operations applied to the target image when training the VAE. The top row corresponds to a real image and the bottom row to a simulated one.}
    \label{fig:vae}
    \vspace*{-6mm}
\end{figure*}

The pixels of a depth image contain a lot of redundant information that is not directly useful for collision avoidance and navigation. Therefore, we encode the depth frames using a convolutional Variational Autoencoder (VAE) \cite{vae}. Its purpose is to map the images coming from the robot to a low-dimensional latent space suitable for collision avoidance and help bridge the reality gap. By passing the data through a bottleneck, the input is compressed to only the most relevant information, namely the geometry's general shape. Additionally, choosing a smaller latent space dimension reduces its capacity to represent finer details. For our task, this leads to a better generalization to unseen objects and is also desirable because we only need to estimate the rough collision shape of objects for obstacle avoidance. Similarly, the VAE removes the input signal's high frequency content and thus filters out noise in the images. This is important since the depth images obtained from the real camera are noisy, see Fig.~\ref{fig:vae}.

To bridge the reality gap, the VAE is trained with images from simulation and the real robot. We define the autoencoder's loss as the reconstruction error between the decoded image and a filtered version of the input image. The filtering operations are inspired by the IP-Basic algorithm \cite{ku2018defense}, which performs simple depth completion. We extend this method to inflate objects and remove their level of detail to extract object bounds rather than exact shapes. As a result, we can mask the main shortcomings of real depth images compared to simulated ones: Missing information, loss of detail, and depth noise. The filter first performs a dilation and hole closing operation. This removes the noise in the planes and along the edges of objects. Next, a dilation with a large kernel is performed on the image but only applied to the remaining black pixels. Finally, a bilateral filter is applied to the resulting image to smooth out shapes. 
These steps are depicted in Fig.~\ref{fig:vae}, which shows how missing information is recovered and the level of detail of objects reduced.

The filtering operations could also be performed as a pre-processing step before passing the image to the VAE. However, this would require running the filtering operations at inference time, which would increase computational load on our mobile hardware. With our method, filtering is implicitly integrated into the VAE forward-pass and therefore incurs no additional computational cost. 

\subsection{Fusing the sequence of images}
\label{sec:lstm}
\begin{figure}
    \centering
    \begin{subfigure}[b]{0.40\columnwidth}
        \centering
        \includegraphics[width=\columnwidth]{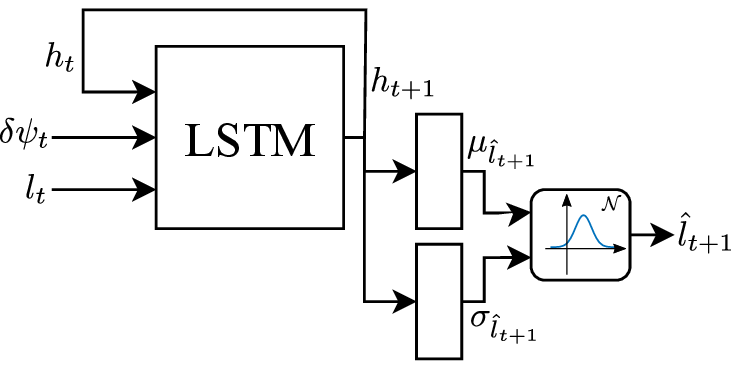}
        \caption{}
        \label{fig:world_model}
    \end{subfigure}%
    ~ \hfill
    \begin{subfigure}[b]{0.59\columnwidth}
        \centering
        \includegraphics[width=\columnwidth]{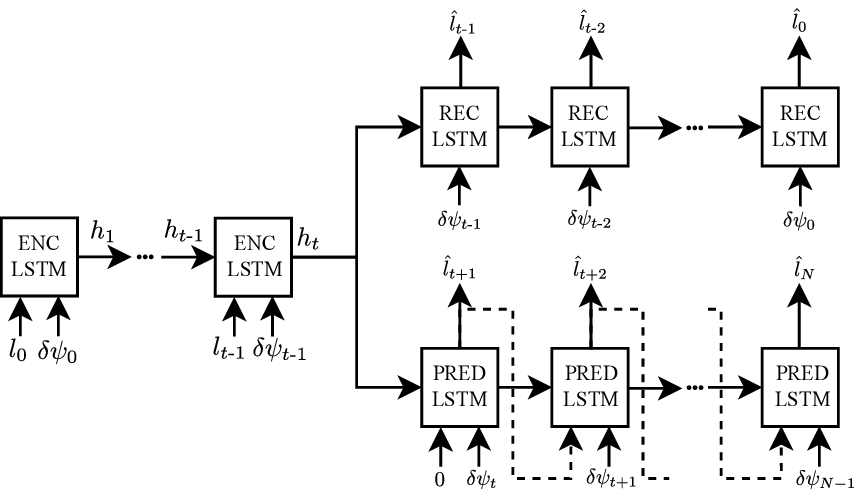}
        \caption{}
        \label{fig:enc_dec_lstm}
    \end{subfigure}
    \caption{(a) Classic approach to learn a representation of the world. Given the belief state $h_t$, a latent $l_t$ and displacement $\delta \psi_{t}$, this network learns a distribution over the next latent $l_{t+1}$  (b) State representation learning with an encoder-decoder LSTM approach. An encoder LSTM computes a belief state $h_t$ from a trajectory, which is then used to predict future latent vectors and reconstruct the past trajectory.}
    \vspace*{-4mm}
\end{figure}

The sequence of depth images $d_{\{0,...,t\}}$ is encoded by the VAE, producing a sequence of latent vectors $l_{\{0,...,t\}}$. 
Along with the camera trajectory $\delta \psi_{\{0,...,t\}}$, it is fed through a Long Short-Term Memory (LSTM)~\cite{lstm} block to produce a hidden state $h_{t+1}$. The hidden state is then mapped using two multilayer perceptrons (MLP) to the estimated mean $\mu_{\hat{l}_{t+1}}$ and standard deviation $\sigma_{\hat{l}_{t+1}}$ of the distribution of the next latent $\l_{t+1}$, which is assumed to be Gaussian (see Fig.~\ref{fig:world_model}). 
We train the model by maximizing the log-likelihood of the measured latent vectors. We also add a regularizing KL-divergence term $D_{KL}(p(\hat{l}_{t+1} | h_{t+1}) || q)$, where $q \sim \mathcal{N}(0, I)$. This makes the predicted distribution consistent with the prior over the latent variables that is introduced in the VAE loss \cite{vae}. 
The signal $h_t$ is a compact representation of the world at time $t$. To predict the next latent, it has to encode the layout of the room, the objects in the scene and their velocity implicitly. 
For the subsequent sim-to-real transfer, the state representation module is trained by mixing trajectories from the simulation and the real robot.

Since the output is a probability distribution that maximizes the measurement log-likelihood, the representation module can be seen as a state estimator similar to a non-linear Kalman filter \cite{kalman}.
Given the current belief state $h_t$, the previously inferred latent $\hat{l}_t$, and input $\delta \psi_{t}$, the next latent $\hat{l}_{t+1}$ can be predicted. This corresponds to the dynamics update, commonly referred to as dreaming~\cite{dreaming} in the learning literature. When it receives a measured latent $l_{t+1}$, a measurement update is performed by inferring the model. Due to the log-likelihood loss, the standard deviation output  $\sigma_{\hat{l}_{t+1}}$ reflects the confidence of the model on how accurately it represents the state of the world. 

In this work, we also compare this method against a more sophisticated sequence to sequence learning approach depicted in Fig.~\ref{fig:enc_dec_lstm}, which is commonly used in natural language processing or video prediction \cite{seq2seq, seq2seq-video}. The encoder LSTM compresses the images and camera trajectory to a representation $h_t$. During training, that representation $h_t$ is fed to a reconstruction LSTM, whose task is to reconstruct the input sequence and a prediction LSTM, whose task is to predict the future. At run-time, only the encoder LSTM is used.

\section{Learning obstacle avoidance}
\label{sec:policy}
In this section, we define the collision avoidance task. The robot is an agent whose task is to navigate in a dynamic environment. It receives a sequence of depth images, the camera's trajectory, a goal to reach, and has to decide on a base velocity command. 
Internally, it uses the approach described in Section \ref{sec:state_representation} to compute the current belief state. The agent will leverage this powerful representation and use it to learn an obstacle-avoiding policy. The policy itself is a simple MLP that receives the representation and the goal as input. 

\subsection{Problem formulation}
We model the problem in simulation and use a model-free RL algorithm. Using a learning-based approach allows us to use the abstract state representation directly, which could not be directly interpreted for use in a model-based setting. However, the use of a simulator requires careful attention to ensure successful sim-to-real transfer. 

Instead of simulating the whole robot and its velocity tracking controller, we simply assume that the robot is made of a single rectangular cuboid whose collision shape is slightly bigger than the robot itself (to have a safety margin when deploying the policy on hardware). The actions are the velocity along the $x$ and $y$ coordinates of the cuboid and its yaw rate, mimicking the effect of the robot's controller. 

This simplified formulation significantly increases simulation speed and is sufficient as long as the robot's controller can track the commanded velocities well enough.

The episode terminates whenever the robot reaches the goal, collides with an obstacle, or a time-out occurs, with respective terminal rewards $r_{obst}$, $r_{goal}$, $r_{to}$. Additionally, we penalize lateral and backward squared velocities, with scaling $\alpha_{lat}$ and $\alpha_{back}$, respectively, since the robot is blind in these directions, and the total traveled distance at the end of the episode,  with scaling $\alpha_{dist}$.

At the beginning of every episode, a new world and goal position are generated, and the camera's tilt, pitch, and attachment position are slightly randomized.

\subsection{Procedural environment generation}
\label{sec:Env}
\begin{figure}
    \centering
    \begin{subfigure}{0.49\columnwidth}
        \centering
        \includegraphics[width=\columnwidth]{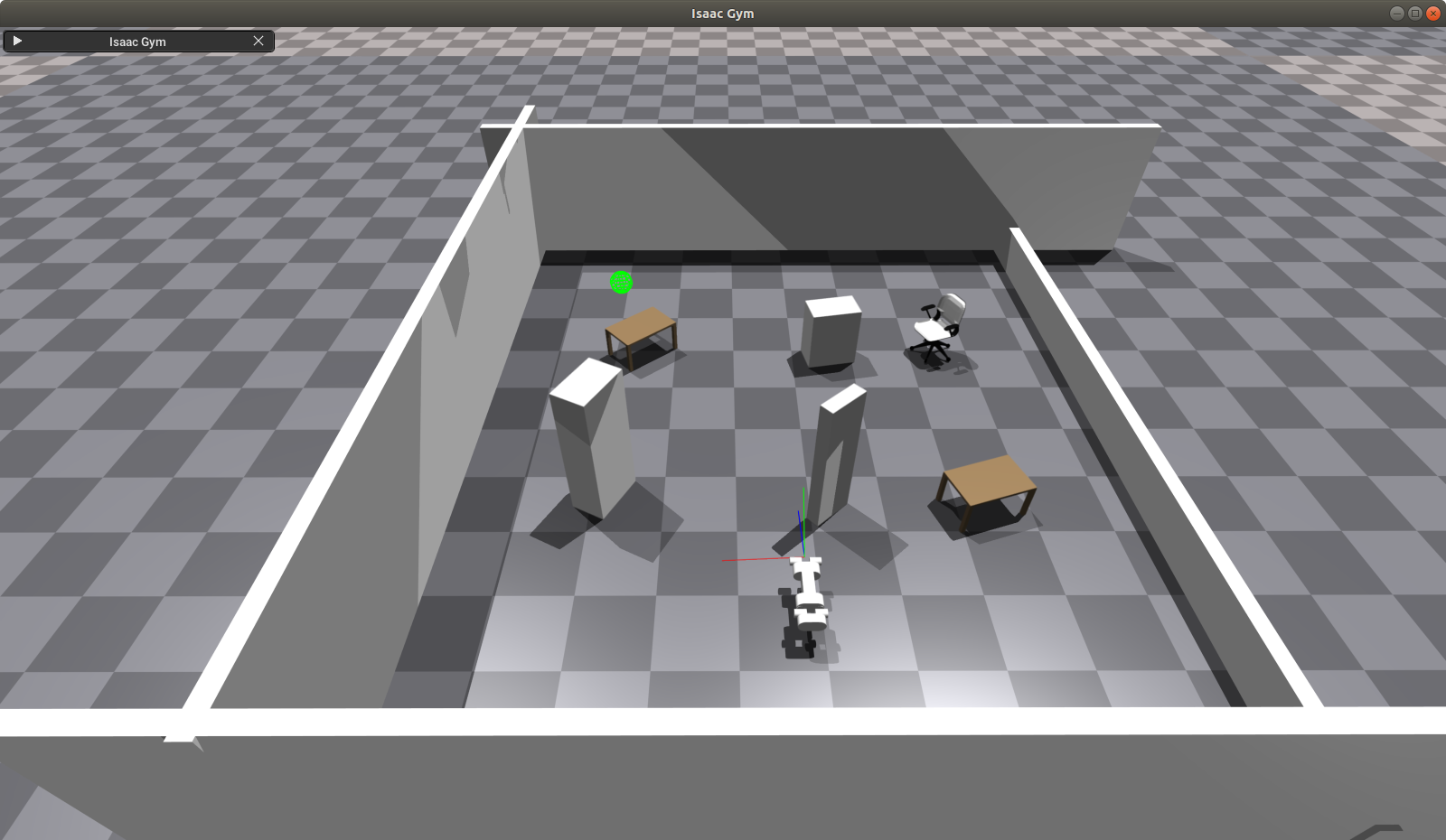}
    \end{subfigure} 
    \begin{subfigure}{0.49\columnwidth}
        \centering
        \includegraphics[width=\columnwidth]{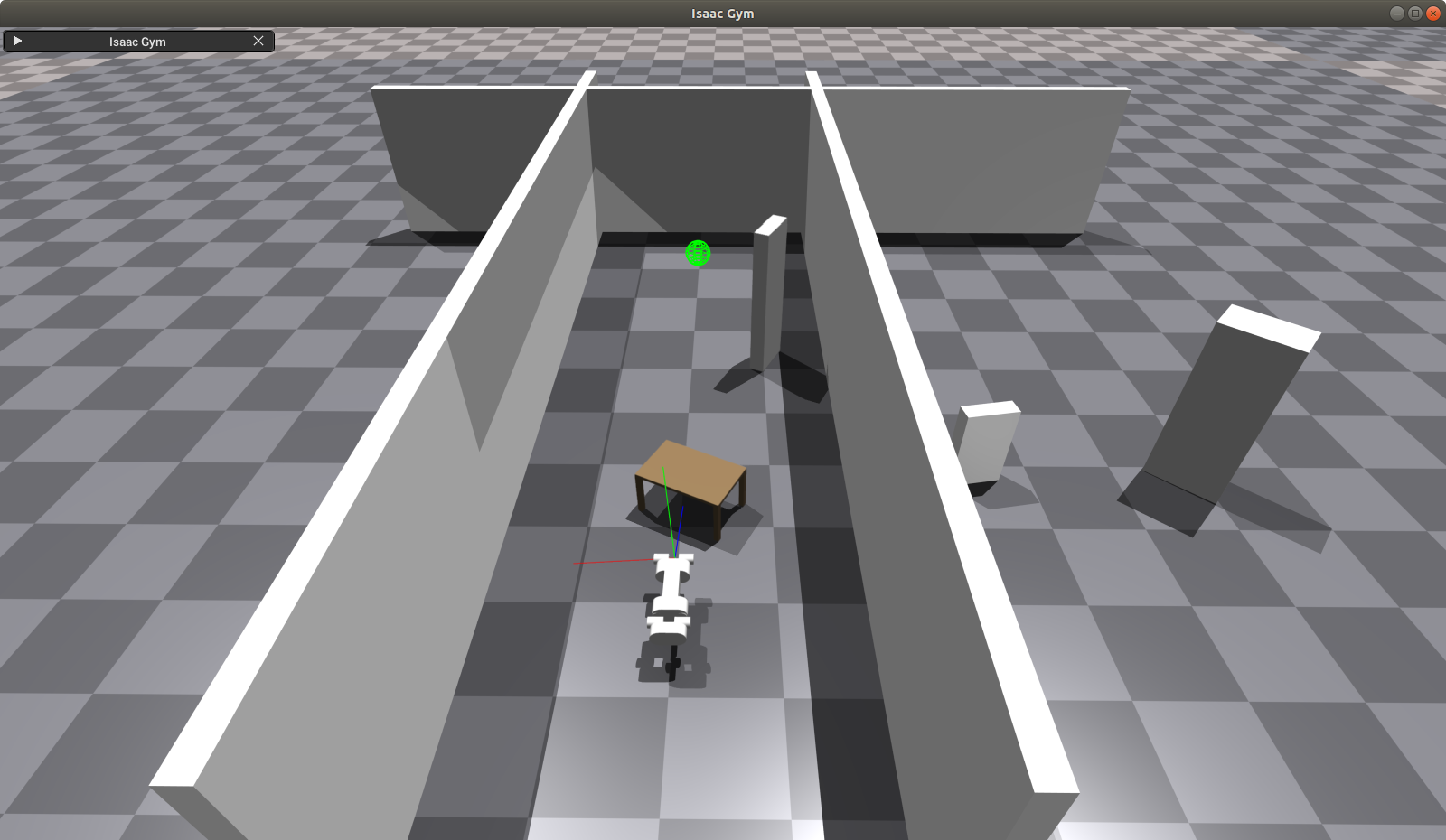}
    \end{subfigure}
    \caption{Typical training scenarios generated in simulation. All of the components in the scene are randomized. The goal point is represented by the green sphere.}
    \label{fig:world}
    \vspace*{-6mm}
\end{figure}

The goal of the environment generator is to create a variety of training cases for the agent and encompass typical scenarios the real robot will face, see Fig.~\ref{fig:world}. Every agent is spawned in its own independent world, which consists of walls and moving obstacles. 

At the beginning of a new episode, the wall heights and the world's narrowness are randomized to simulate narrow corridors up to open worlds. We also randomize the object categories, their position, and velocity.

To control the difficulty of the task, the maximum velocity of objects and the object density in the scene can be modified, which can be exploited for curriculum learning.

\subsection{Bringing all the components together}
The whole pipeline contains a lot of interdependent components. We train using the following sequence: 
\begin{enumerate}
    \item We first collect images to train the VAE by spawning the camera at random locations in random environments in simulation and from different real-world deployments of the robot. We use around 50'000 images from simulation and 20'000 images from reality.
    \item A baseline policy, which uses the VAE's latent vector as input is trained in a static world. 
    \item We collect trajectories by rolling out the baseline policy in dynamic environments and train the LSTM and the MLPs with the VAE frozen. Despite failing quite often in such environments, we found it is sufficient to train the model. We collect around 500'000 transitions which corresponds to 14 hours of real time experience. We also use 3 hours of trajectories collected on the real robot. 
    \item  After the representation model is trained, we train the final policy using the belief state as input in randomly generated dynamic environments in simulation.
    \item The resulting policy can be deployed on the real robot.
\end{enumerate}

\section{Experimental Results}
\label{sec:result}
We experimentally evaluate our approach in simulation and on the real robot in different scenarios, compare the approach to other baselines, and assess the importance of each component of the pipeline. We use NVIDIA's Isaac Gym\footnote{\href{https://developer.nvidia.com/isaac-gym}{https://developer.nvidia.com/isaac-gym}} environment to generate the synthetic data and train the policy\footnote{The training hyper-parameters are described in Appendix~\ref{app:parameters}.}. The depth images have a dimension of $128\times256$.

\subsection{State representation}
\label{sec:kalman}
In this Section, we showcase the state representation learner, especially in the state estimation setting. All the trajectories shown are taken from validation data.

\begin{figure*}[t!]
    \vspace*{5mm}
    \centering
    \captionsetup{singlelinecheck=off}
    \setlength\lineskip{4pt}
    \hspace*{\fill}
    \begin{subfigure}[t]{0.15\textwidth}
        \centering
        \includegraphics[width=\textwidth]{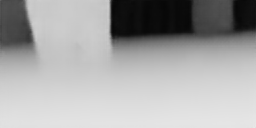}
    \end{subfigure} \hfill
    \begin{subfigure}[t]{0.15\textwidth}
        \centering
        \includegraphics[width=\textwidth]{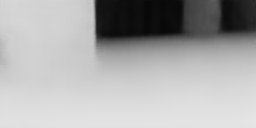}
    \end{subfigure} \hfill
    \begin{subfigure}[t]{0.15\textwidth}
        \centering
        \includegraphics[width=\textwidth]{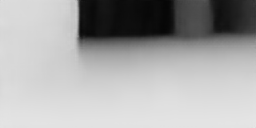}
    \end{subfigure} \hfill
    \hspace{.03\textwidth}
    \begin{subfigure}[t]{0.15\textwidth}
        \centering
        \includegraphics[width=\textwidth]{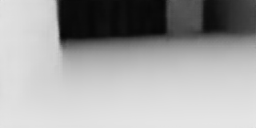}
    \end{subfigure} \hfill
    \begin{subfigure}[t]{0.15\textwidth}
        \centering
        \includegraphics[width=\textwidth]{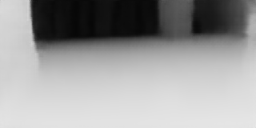}
    \end{subfigure} \hfill
    \begin{subfigure}[t]{0.15\textwidth}
        \centering
        \includegraphics[width=\textwidth]{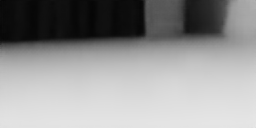}
    \end{subfigure}    
    \hspace*{\fill}
    
    \hspace*{\fill}
    \begin{subfigure}[t]{0.15\textwidth}
        \centering
        \includegraphics[width=\textwidth]{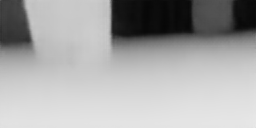}
        \caption{\centering $t=0.0s$ \newline $\sigma=0.03$}
    \end{subfigure} \hfill
    \begin{subfigure}[t]{0.15\textwidth}
        \centering
        \includegraphics[width=\textwidth]{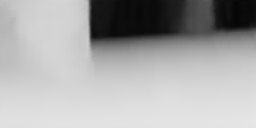}
        \caption{\centering $t=0.5s$ \newline $\sigma=0.07$}
    \end{subfigure} \hfill
    \begin{subfigure}[t]{0.15\textwidth}
        \centering
        \includegraphics[width=\textwidth]{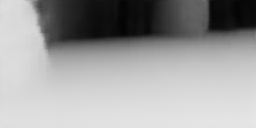}
        \caption{\centering $t=1.0s$ \newline $\sigma=0.23$}
    \end{subfigure} \hfill
    \hspace{.03\textwidth}
    \begin{subfigure}[t]{0.15\textwidth}
        \centering
        \includegraphics[width=\textwidth]{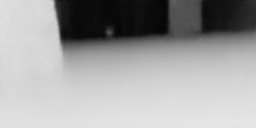}
        \caption{\centering $t=1.5s$ \newline $\sigma=0.11$}
    \end{subfigure} \hfill
    \begin{subfigure}[t]{0.15\textwidth}
        \centering
        \includegraphics[width=\textwidth]{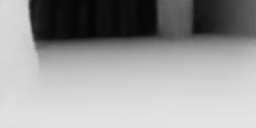}
        \caption{\centering $t=2.0s$ \newline $\sigma=0.07$}
    \end{subfigure} \hfill
    \begin{subfigure}[t]{0.15\textwidth}
        \centering
        \includegraphics[width=\textwidth]{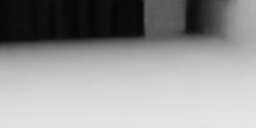}
        \caption{\centering $t=2.5s$ \newline $\sigma=0.06$}
    \end{subfigure}    
    \hspace*{\fill}
    \caption{State estimation along a trajectory. The top row represents the decoded latent coming from ground truth images. The bottom row is the decoded latent coming from the state estimator. Up to time $t=1.0s$, the model is dreaming, which corresponds to the dynamics update of our system, see Section \ref{sec:lstm}. Then, until time $t=2.5s$, the state estimator receives depth images, corresponding to a measurement update. Note that we only show every fifth update here.}
    \label{fig:dream}
    \vspace*{-5mm}
\end{figure*}

Fig.~\ref{fig:dream} depicts an example trajectory. To obtain this sequence, we first warm-start ($t<0$) the network by feeding ground truth data. From time $t=0s$ to $t=1.0s$, we perform several dynamics updates (dreaming). Then, the network receives measurements again until $t=2.5s$. As mentioned in Section \ref{sec:kalman}, the network's standard deviation output will represent the confidence of the network in the accuracy of its belief state. It can be seen that the standard deviation increases while dreaming and decreases again when receiving measurements. This property arises because the further into the future it dreams, the more errors accumulate, leading to blurred images. Therefore, to keep the loss low, the standard deviation must increase to encompass all potential scenarios that could occur. This valuable information could be used at a later stage as input to the policy, or to scale its output appropriately.

Along a trajectory, when an object leaves the field of view, the network cannot reconstruct it properly when it re-enters the field-of-view in the dream. This is likely due to the training data distribution, where successive frames dominate the loss, rather than the rare occasions where objects reappear multiple time steps ahead.

\subsection{Obstacle avoidance using a representation}
\label{sec:sim}
We evaluate the performance of the obstacle-avoiding policy based on three different state representations in simulation. The first representation is based on the latent space of the VAE, the second on the hidden state of the LSTM described in Fig.~\ref{fig:world_model}, and the last one on the hidden state of the encoder LSTM described in Fig.~\ref{fig:enc_dec_lstm}. We refer to these as reactive, world model, and seq-to-seq, respectively. The representation vectors are directly given as input to the MLP policy. The reactive formulation has no notion of time and can only solve the task by displaying reactive behavior.



We assess the pipeline's spatial awareness and evaluate how well it copes with dynamic obstacles.

\begin{figure}
    \centering
    \begin{subfigure}[b]{\columnwidth}
        \centering
        \includegraphics[width=0.49\columnwidth]{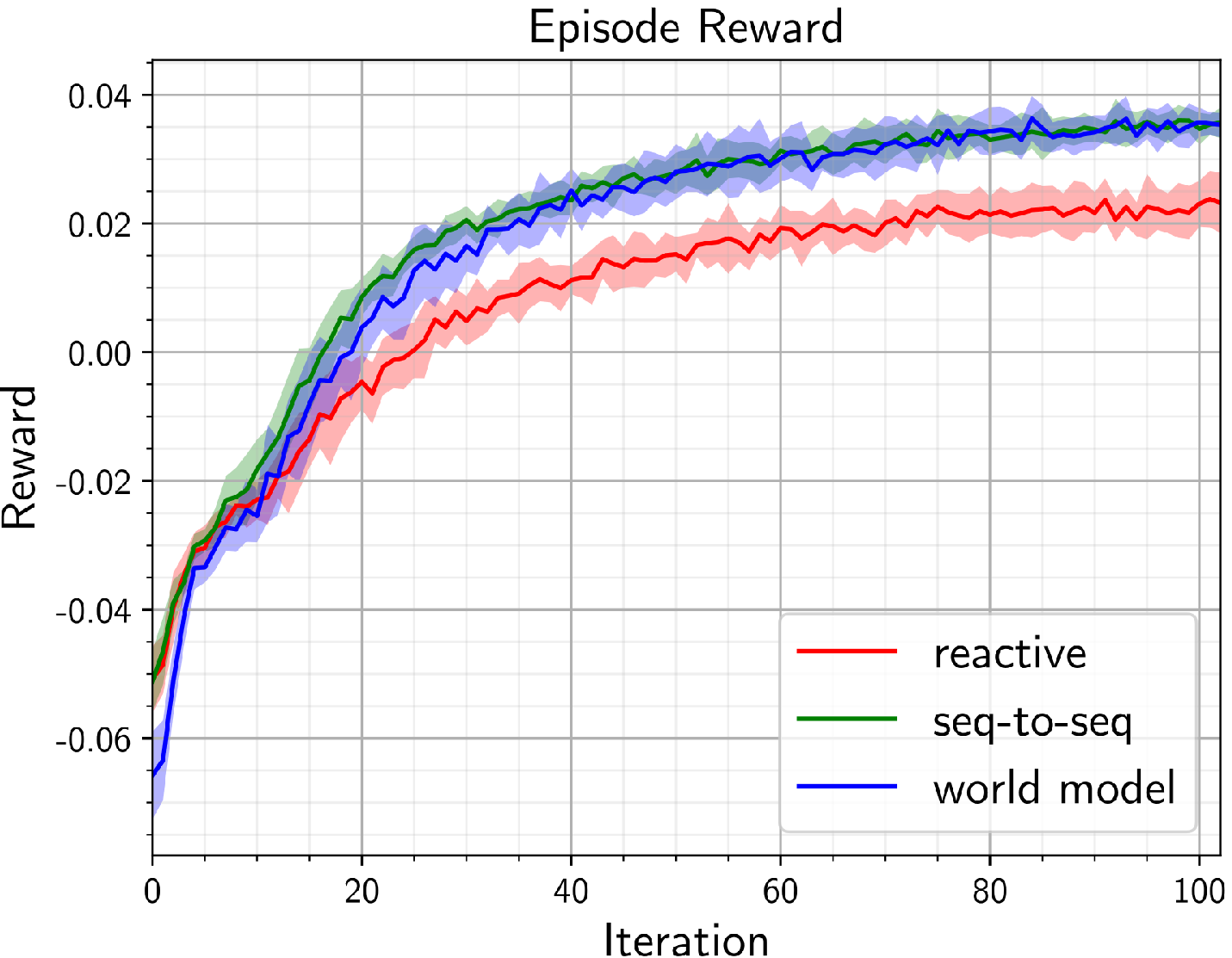}
        \includegraphics[width=0.49\columnwidth]{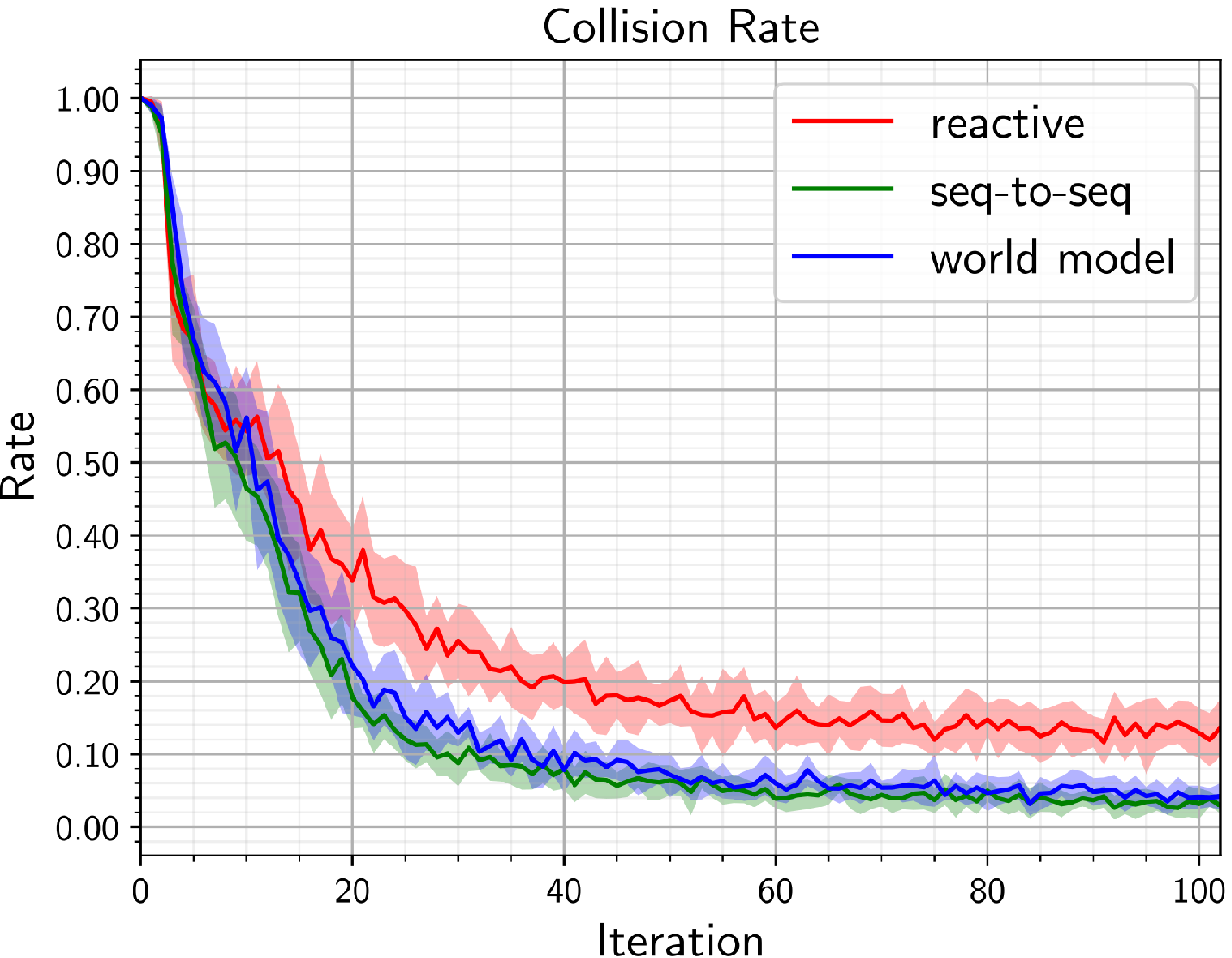}
        \caption{Cluttered static scenes.}
        \vspace*{1mm}
        \label{fig:static}
     \end{subfigure}
    \begin{subfigure}[b]{\columnwidth}
        \centering
        \includegraphics[width=0.49\columnwidth]{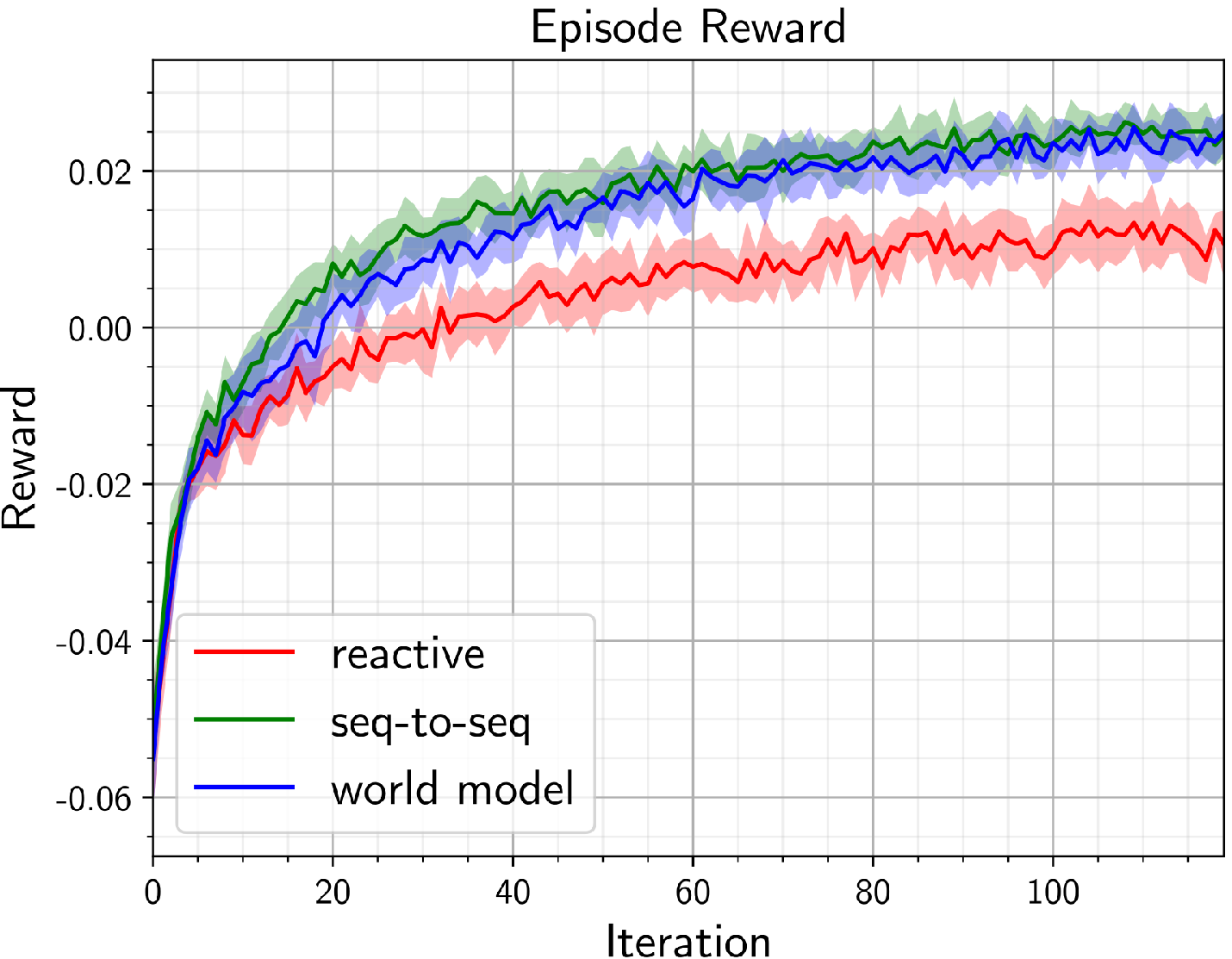}
        \includegraphics[width=0.49\columnwidth]{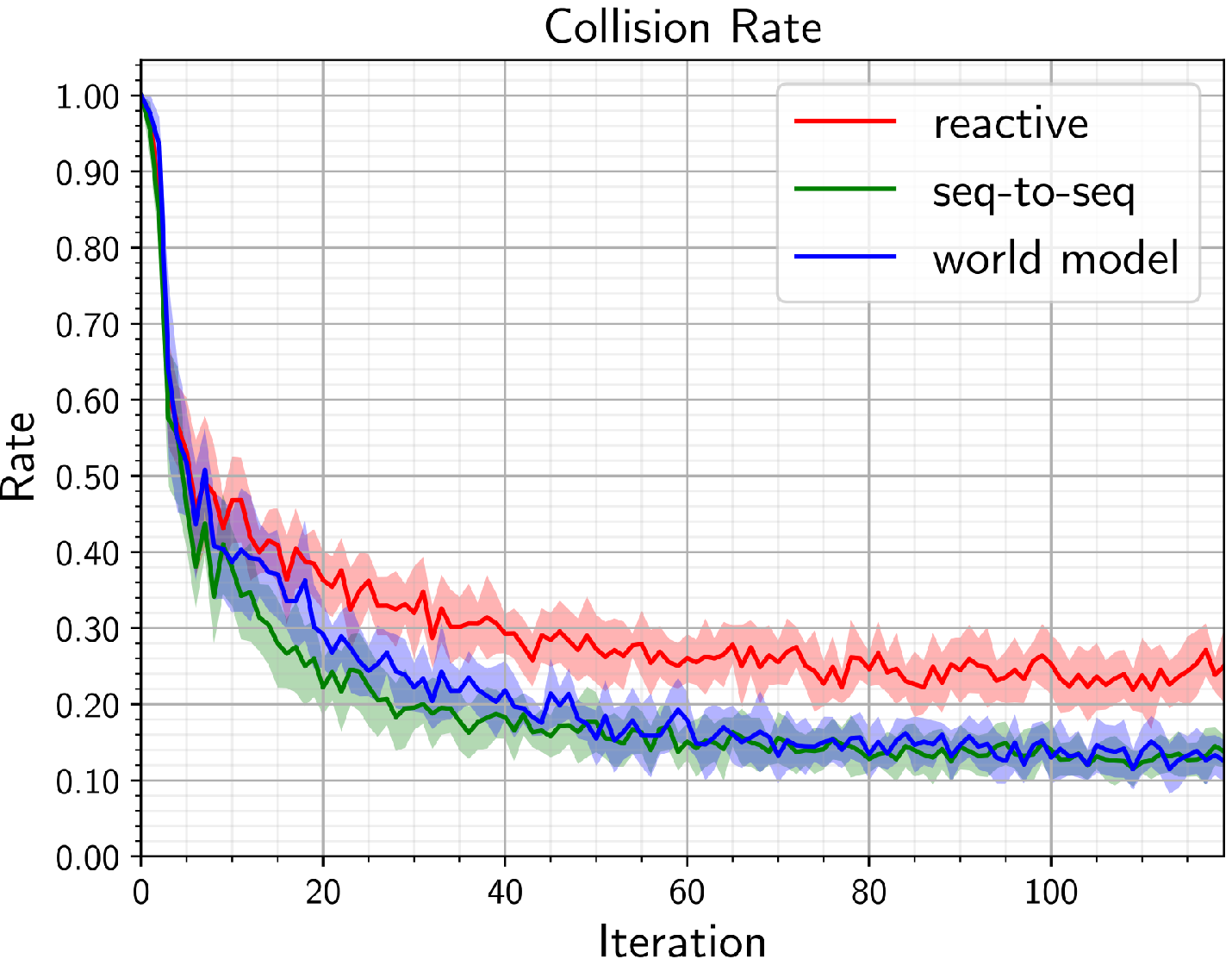}
        \caption{Dynamic scenes.}
        \label{fig:dyn}
    \end{subfigure}
    \caption{Performance of the different formulations in static and dynamic scenes. The values are obtained by running the polices on 1000 unseen environments (random layout and object velocities) at each iteration of the policy learning stage.}
    \vspace*{-8mm}
\end{figure}

\paragraph{Cluttered static environment}
\label{sec:clutter}
In this experiment, we evaluate the different policy formulations in a highly cluttered static scene, see Fig.~\ref{fig:static}. The world model and seq-to-seq policies achieve a better performance than the reactive policy. They reach a rate of 3\% at full convergence, while the reactive policy converges towards 13\%. 
On average, the reactive policy also takes a longer path to reach the goal, while the memory-based policies dedicate less control effort overall and move faster towards the goal. 


The main limitation of the reactive policy is that it can only react to the information available at the current time step. Therefore, when the policy fails, it usually first successfully goes around an obstacle but then collides with it when it leaves the field of view. On the contrary, this rarely happens with the world model or the seq-to-seq policies. When they collide with an obstacle, it is mostly due to the noise added to the robot's actions to make the policy more robust on the real robot. 

 These experiments show that the memory-enabled policies have a better understanding of the robot's immediate surroundings and the scene's spatial configuration. Consequently, they can use less control effort and complete the navigation task more quickly.

\paragraph{Dynamic environments} 
In this experiment, we evaluate the different policy formulations in a dynamic scene. The test scenes are less cluttered (10\% lower object spawn rate) than in the previous experiments because it would otherwise result in too difficult scenarios. The policies have to solve a more complicated task, since they need to display spatial awareness and show foresight and understand the evolution of the scene in time. Therefore, the behavior converges after more iterations and reaches a higher collision rate (25\% for the reactive and 13\% for the other two policies),  see Fig.~\ref{fig:dyn}. 


Despite having no memory, the reactive policy is still to some extent capable of avoiding oncoming objects. However, this comes at the cost of control effort. Indeed, the reactive policy uses twice as much velocity in backward direction along a trajectory. This is because it does not know the velocity of an oncoming obstacle. It has to be particularly cautious and needs to control for the worst-case scenario. On the other hand, the world model and seq-to-seq policies choose a shorter path to the goal and display lower control effort. 

The higher failure rates of the different policies compared to the static scenario can be explained by two contributing factors. First, following the observation made in \ref{sec:kalman}, the state estimator is prone to forgetting objects that have left the field of view. The consequence of this effect is further aggravated when the object in question is moving towards the robot. Second, just using the hidden state at the current time step might not sufficient for navigation. Ideally, the policy should be given an idea of what the configuration of the scene will look like a couple of seconds ahead. The policy could, for example, be given a dreamed version of the scene in the future. This is left for future work. 

These experiments show that the RL trained policies can exploit the additional information brought by the belief state of the representation learner. The world model and seq-to-seq policies display very similar performance, even though the sequence to sequence approach is trained to encode more information.

\subsection{Comparison with baselines}

To highlight the representation learner's effectiveness, we compare our seq-to-seq policy with three end-to-end baselines in simulation. In the first one, denoted by \textit{end-to-end CNN}, the policy directly takes the depth images as input and consists of a convolutional neural network (CNN) followed by an MLP, which also takes the goal as input. In the second one, denoted by \textit{end-to-end CNN+LSTM}, the CNN's output is fed through an LSTM along with the goal and the resulting hidden state given to an MLP. This recurrent policy is the end-to-end counterpart to the world model policy. We also report the performance of a blind MLP policy that only receives the goal as input to reflect the difficulty of the task. In these experiments, the agent has to reach the goal within 6 seconds without collision in an unseen room layout.



The results shown in Table~\ref{tab:comparison} indicate that using the state representation results not only in a more sample efficient RL stage but also yields a superior policy. Moreover, for the end-to-end approaches, the RL algorithm has to keep track of the images along the trajectories, while our method only needs the low dimensional representation vector in our method. This results in a 2 to 3 times faster computation time per learning iteration. We also found that the end-to-end approaches are more sensitive to hyperparameters such as the learning rate and require a curriculum where the number of objects in the scene is gradually increased to converge. 
However, the main limitation of the end-to-end approaches is that the networks never receive data from the real world during training and are therefore prone to fail when deployed on the real robot, as we show in our experiment in Section~\ref{sec:simtoreal}. A solution would be to train these policies directly on the real robot, but setting up such an obstacle avoidance experiment would be very time-consuming and cumbersome.



\begin{table*}[t]
    \vspace*{2mm}
    \centering
    \caption{Comparison of our method against a blind and end-to-end baselines. Our policy reaches a better performance with fewer samples, in less time, and does not require a curriculum to converge. We put the training time of the state representation module in parenthesis. This module is frozen during the policy learning stage.}
    \begin{tabular}{c | c c c c c c c c}
        \hline
        {Experiment} & {Policy} & \makecell{Failure \\ Rate [$\%$]} & \makecell{Traveled \\ Distance [$m$]} & \makecell{Reward} & \makecell{Iterations to  \\ convergence} & \makecell{Time to \\ convergence [$\si{min}$]} & \makecell{Requires \\ Curriculum}  & \makecell{Successful \\ sim-to-real}\\
        \hline
        \multirow{4}*{\textbf{Static}} & {Ours} & 3 & 3.50 & 0.035 & 100 & 10 (30) & No & \cmark \\
                                       & {Blind} & 75 & 3.90 & -0.030 & 50 & 5 & No & \xmark \\
                                       & {End-to-end CNN} & 10 & 3.55 & 0.022 & 240 & 64 & Yes & \xmark\\
                                       & {End-to-end CNN+LSTM} & 6 & 3.45 & 0.025 & 280 & 93 & Yes & \xmark\\
        \hline
        \multirow{4}*{\textbf{Dynamic}} & {Ours} & 11 & 3.78 & 0.025 & 120 & 12 (30) & No & \cmark \\
                                        & {Blind} & 76 & 4.00 & -0.040 & 60 & 6 & No & \xmark\\
                                        & {End-to-end CNN} & 20 & 3.75 & 0.015 & 260 & 70 & Yes & \xmark\\
                                        & {End-to-end CNN+LSTM} & 15 & 3.68 & 0.020 & 280 & 93 & Yes & \xmark\\
        \hline
    \end{tabular}
    \label{tab:comparison}
    \vspace*{-3mm}
\end{table*}

\subsection{Sim-to-Real transfer}
\label{sec:simtoreal}
We deploy the seq-to-seq policy on the quadrupedal robot ANYmal \cite{anymal} equipped with a Realsense D435 camera. All the computations take place on the GPU of the Jetson AGX Xavier on-board computer. After updating the belief state and computing a new policy output, the command is sent to a velocity tracking controller \cite{8260889}. 

To help overcome the reality gap between velocity tracking performance in simulation and on the real system, we scale the policy's velocity commands that are sent to the velocity tracking controller. These scaling factors are tuned manually to obtain the best performance on the robot. Otherwise, the transfer is straightforward, and no further tuning was required to make the policy work in real life. 

\begin{figure}[t]
\vspace*{2mm}
    \centering
    \includegraphics[width=1.0\columnwidth]{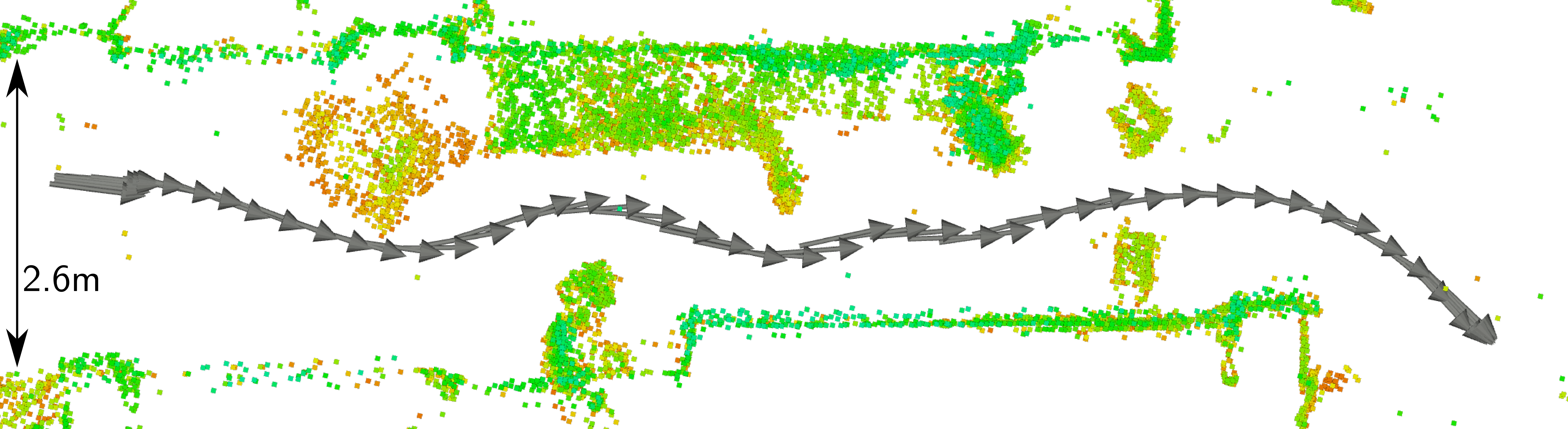}
    \caption{Path of the robot in a cluttered environment. The arrows represent the heading of the robot and the dots the point cloud of the narrow passage.}
    \label{fig:nav}
    \vspace*{-5mm}
\end{figure}

\paragraph{Navigating in a static environment}
First, we let the robot walk in a cluttered static lab environment towards a goal point. The robot cannot reach the target in a straight line. As can be seen in Fig.~\ref{fig:nav}, the robot can successfully navigate this environment. It is full of shapes that have not been seen during training, meaning that the state representation module and the policy both generalize to unseen environments. We attribute this to the filtering operation carried out implicitly by the VAE, which removes details of objects. Also, we inadvertently deployed the robot once with an LSTM that has seen no real data during training. The policy still worked on the robot, which indicates that the latent space produced by the VAE is very effective for sim-to-real transfer and that latent space dynamics do not differ between simulation and reality.
The robot very rarely collides with obstacles in the field of view. However, when the camera is tilted away from an obstacle or a wall that was previously seen by the camera, it might still collide with it. This is again consistent with the observation made in \ref{sec:kalman} and shown in the previous experiments. This effect is further amplified when using the real robot. As mentioned before, this could be tackled by adding more cameras to the pipeline.

\paragraph{Avoiding dynamic obstacles}
Next we let the robot walk towards a goal in a dynamic scene indoors and outdoors. Along the path, a person walks in front of the robot in various directions and actively tries to block the path.
The robot clearly reacts to oncoming obstacles in a quick manner. 

Nevertheless, the robot stops walking when a person approaches and then goes around. It will also deliberately go in the opposite direction of the goal to avoid an oncoming obstacle.
The robot can avoid a collision even when an obstacle enters the camera field-of-view from the side at a close distance. Please refer to the video\footnote{\href{https://youtu.be/CICWcLJ3aPs}{https://youtu.be/CICWcLJ3aPs}} to see the emerging behaviors. 

We also attempted to transfer the end-to-end policies using filtered depth images. The robot is only partially able to avoid obstacles and never manages to reach the goal. As mentioned before, this is due to the fact that these policies never see real data during training.

\vspace*{-2mm}
\subsection{Ablation study}
\label{sec:ablation}
In this ablation study, we assess the importance of using real data when training the state representation and the importance of the filtering step described in Section \ref{sec:VAE}.

In a first step, we train the representation module once with and once without the filtering step. Then, we train the corresponding policies in a static scene that contains new objects that have not been seen before. The policy with filtering converges to a collision rate of 8\%, while the one without to 30\%.

Next, we train the representation module without filtering and deploy the corresponding policy on the real robot. Despite reaching a low collision rate in simulation, the policy behaves erratically and cannot avoid obstacles or reach the goal.

Finally, we train the representation module with the filtering step but without using real data. Similarly, the policy works in simulation but fails on the real robot. 

In conclusion, filtering and mixing real data are key components to make the sim to real transfer successful. The filtering step also improves the representation module's generalization performance to unseen scenarios.




\section{Conclusion}
This work presents a successful sim-to-real transfer of a learning-based navigation pipeline for cluttered and dynamic scenes. A network to estimate a time-dependent model of the world is trained through unsupervised learning, the output of which is used to train a navigation policy in simulation using RL. The pipeline can operate indoors and outdoors in scenes that have not been seen during training.\\
In future work, we will extend this work to operate in rough terrain by designing more complex environments with objects moving along the terrain. We will also work on improving the shortsightedness of the representation network. 


\section*{APPENDIX}
\label{app:parameters}
\subsection{Experiments} In the experiments, the robot has a maximum $x$ and $y$ velocity of 1 $m/s$ and a maximum yaw rate of 1 $rad/s$. The velocities of the objects in the scene are sampled uniformly with a maximum velocity of 0.8 $m/s$.
\vspace*{-2mm}
\subsection{State representation} The VAE has a latent space dimension of 32. The LSTM networks have 2 layers and a hidden state of dimension 128, which is mapped to the parameters of the latent distribution using two 1 layer MLPs with a linear activation. We train the networks using the Adam \cite{adam} optimizer with a learning rate of 1e-3. 
\vspace*{-2mm}
\subsection{Policy learning} We train the policies using the Proximal Policy Optimization \cite{PPO} algorithm. The policy and the value function are simple MLPs with tanh activations. During training, we run 100 environments in parallel and collect 60 transitions per learning iteration per environment and use the Adam \cite{adam} optimizer with a learning rate of 1e-3. 



For the reward parameters described in Section~\ref{sec:policy}, we use the following values for all our experiments:
\begin{center}
\begin{tabular}{ l|l l l l l l } 
 Term & $r_{obst}$ & $r_{goal}$ & $r_{to}$ & $\alpha_{lat}$ &  $\alpha_{back}$ & $\alpha_{dist}$ \\ 
  \hline 
 Value & -2.0 & 2.0 & 0.0 & -0.05 & -0.005 & -0.05
\end{tabular}
\end{center}

\bibliographystyle{IEEEtran}
\bibliography{IEEEabrv,root}

\begin{thebibliography}{10}
\providecommand{\url}[1]{#1}
\csname url@rmstyle\endcsname
\providecommand{\newblock}{\relax}
\providecommand{\bibinfo}[2]{#2}
\providecommand\BIBentrySTDinterwordspacing{\spaceskip=0pt\relax}
\providecommand\BIBentryALTinterwordstretchfactor{4}
\providecommand\BIBentryALTinterwordspacing{\spaceskip=\fontdimen2\font plus
\BIBentryALTinterwordstretchfactor\fontdimen3\font minus
  \fontdimen4\font\relax}
\providecommand\BIBforeignlanguage[2]{{%
\expandafter\ifx\csname l@#1\endcsname\relax
\typeout{** WARNING: IEEEtran.bst: No hyphenation pattern has been}%
\typeout{** loaded for the language `#1'. Using the pattern for}%
\typeout{** the default language instead.}%
\else
\language=\csname l@#1\endcsname
\fi
#2}}

\bibitem{4621214}
D.~{Ferguson}, M.~{Darms}, C.~{Urmson}, and S.~{Kolski}, ``Detection,
  prediction, and avoidance of dynamic obstacles in urban environments,'' in
  \emph{2008 IEEE Intelligent Vehicles Symposium}, 2008, pp. 1149--1154.

\bibitem{dwa}
D.~Fox, W.~Burgard, and S.~Thrun, ``The dynamic window approach to collision
  avoidance,'' \emph{Robotics and Automation Magazine, IEEE}, vol.~4, pp. 23 --
  33, 04 1997.

\bibitem{7743420}
A.~C. {Woods}, H.~M. {Lay}, and Q.~P. {Ha}, ``A novel extended potential field
  controller for use on aerial robots,'' in \emph{2016 IEEE International
  Conference on Automation Science and Engineering (CASE)}, 2016, pp. 286--291.

\bibitem{44033}
J.~{Borenstein} and Y.~{Koren}, ``Real-time obstacle avoidance for fast mobile
  robots,'' \emph{IEEE Transactions on Systems, Man, and Cybernetics}, vol.~19,
  no.~5, pp. 1179--1187, 1989.

\bibitem{obstacle}
T.~Eppenberger, G.~Cesari, M.~Dymczyk, R.~Siegwart, and R.~Dube, ``Leveraging
  stereo-camera data for real-time dynamic obstacle detection and tracking,''
  07 2020.

\bibitem{lillicrap2016RL}
T.~P. Lillicrap, J.~J. Hunt, A.~Pritzel, N.~Heess, T.~Erez, Y.~Tassa,
  D.~Silver, and D.~Wierstra, ``Continuous control with deep reinforcement
  learning,'' in \emph{4th International Conference on Learning
  Representations, {ICLR} 2016, Conference Track Proceedings}, 2016.

\bibitem{hwangbo2019RL}
J.~Hwangbo, J.~Lee, A.~Dosovitskiy, D.~Bellicoso, V.~Tsounis, V.~Koltun, and
  M.~Hutter, ``Learning agile and dynamic motor skills for legged robots,''
  \emph{Science Robotics}, vol.~4, no.~26, 2019.

\bibitem{SadeghiL17}
F.~Sadeghi and S.~Levine, ``{CAD2RL:} real single-image flight without a single
  real image,'' in \emph{Robotics: Science and Systems XIII, Massachusetts
  Institute of Technology, Cambridge, Massachusetts, USA}, 2017.

\bibitem{pfeiffer2017perception}
M.~Pfeiffer, M.~Schaeuble, J.~Nieto, R.~Siegwart, and C.~Cadena, ``From
  perception to decision: A data-driven approach to end-to-end motion planning
  for autonomous ground robots,'' in \emph{IEEE International Conference on
  Robotics and Automation (ICRA)}.\hskip 1em plus 0.5em minus 0.4em\relax IEEE,
  2017, p. 1527–1533.

\bibitem{tai2018social}
L.~Tai, J.~Zhang, M.~Liu, and W.~Burgard, ``Socially compliant navigation
  through raw depth inputs with generative adversarial imitation learning,'' in
  \emph{2018 IEEE International Conference on Robotics and Automation (ICRA)},
  2018, pp. 1111--1117.

\bibitem{8461113}
P.~{Long}, T.~{Fan}, X.~{Liao}, W.~{Liu}, H.~{Zhang}, and J.~{Pan}, ``Towards
  optimally decentralized multi-robot collision avoidance via deep
  reinforcement learning,'' in \emph{2018 IEEE International Conference on
  Robotics and Automation (ICRA)}, 2018, pp. 6252--6259.

\bibitem{kaufmann2020RSS}
K.~Elia, L.~Antonio, R.~René, M.~Matthias, K.~Vladlen, and S.~Davide, ``Deep
  drone acrobatics,'' \emph{RSS: Robotics, Science, and Systems}, 2020.

\bibitem{Tobin2017DomainRF}
J.~Tobin, R.~H. Fong, A.~Ray, J.~Schneider, W.~Zaremba, and P.~Abbeel, ``Domain
  randomization for transferring deep neural networks from simulation to the
  real world,'' \emph{2017 IEEE/RSJ International Conference on Intelligent
  Robots and Systems (IROS)}, pp. 23--30, 2017.

\bibitem{anymal}
M.~Hutter, C.~Gehring, D.~Jud, \emph{et~al.}, ``Anymal - a highly mobile and
  dynamic quadrupedal robot,'' in \emph{2016 IEEE/RSJ International Conference
  on Intelligent Robots and Systems (IROS)}.\hskip 1em plus 0.5em minus
  0.4em\relax IEEE, 2016, pp. 38 -- 44.

\bibitem{SFnav}
J.~Michels, A.~Saxena, and A.~Y. Ng, ``High speed obstacle avoidance using
  monocular vision and reinforcement learning,'' in \emph{Proceedings of the
  22Nd International Conference on Machine Learning}, ser. ICML '05.\hskip 1em
  plus 0.5em minus 0.4em\relax ACM, 2005, pp. 593--600.

\bibitem{Kaufmann2018DeepDR}
E.~Kaufmann, A.~Loquercio, R.~Ranftl, A.~Dosovitskiy, V.~Koltun, and
  D.~Scaramuzza, ``Deep drone racing: Learning agile flight in dynamic
  environments,'' in \emph{CoRL}, 2018.

\bibitem{DroNet}
A.~{Loquercio}, A.~I. {Maqueda}, C.~R. {del-Blanco}, and D.~{Scaramuzza},
  ``Dronet: Learning to fly by driving,'' \emph{IEEE Robotics and Automation
  Letters}, vol.~3, no.~2, pp. 1088--1095, 2018.

\bibitem{forestIdsia}
A.~Giusti, J.~Guzzi, D.~C.~Ciresan, \emph{et~al.}, ``A machine learning
  approach to visual perception of forest trails for mobile robots,''
  \emph{IEEE Robotics and Automation Letters}, vol.~1, pp. 1--1, 01 2015.

\bibitem{xie2017towards}
L.~Xie, S.~Wang, A.~Markham, and N.~Trigoni, ``Towards monocular vision based
  obstacle avoidance through deep reinforcement learning,'' in \emph{Robotics:
  Science and Systems Workshop 2017: New Frontiers for Deep Learning in
  Robotics}, 2017.

\bibitem{Everett2018MotionPA}
M.~Everett, Y.~F. Chen, and J.~P. How, ``Motion planning among dynamic,
  decision-making agents with deep reinforcement learning,'' \emph{2018
  IEEE/RSJ International Conference on Intelligent Robots and Systems (IROS)},
  pp. 3052--3059, 2018.

\bibitem{Badgr}
G.~Kahn, P.~Abbeel, and S.~Levine, ``{BADGR:} an autonomous self-supervised
  learning-based navigation system,'' \emph{CoRR}, vol. abs/2002.05700, 2020.

\bibitem{visNavOC}
S.~Bansal, V.~Tolani, S.~Gupta, J.~Malik, and C.~Tomlin, ``Combining optimal
  control and learning for visual navigation in novel environments,'' ser.
  Proceedings of Machine Learning Research, vol. 100.\hskip 1em plus 0.5em
  minus 0.4em\relax PMLR, 2020.

\bibitem{mueller2018driving}
M.~Müller, A.~Dosovitskiy, B.~Ghanem, and V.~Koltun, ``Driving policy transfer
  via modularity and abstraction,'' \emph{CoRL}, 2018.

\bibitem{Sadeghi19}
F.~Sadeghi, ``Divis: Domain invariant visual servoing for collision-free goal
  reaching,'' in \emph{Robotics: Science and Systems XV, Freiburg im Breisgau,
  Germany}, A.~Bicchi, H.~Kress{-}Gazit, and S.~Hutchinson, Eds., 2019.

\bibitem{AutoRL}
H.-T.~L. Chiang, A.~Faust, M.~Fiser, and A.~Francis, ``Learning navigation
  behaviors end-to-end with autorl,'' \emph{IEEE Robotics and Automation
  Letters}, vol.~PP, 2019.

\bibitem{pedestrianNav}
T.~Fan, X.~Cheng, J.~Pan, P.~Long, W.~Liu, R.~Yang, and D.~Manocha, ``Getting
  robots unfrozen and unlost in dense pedestrian crowds,'' \emph{IEEE Robotics
  and Automation Letters}, vol.~PP, 2019.

\bibitem{LESORT2018379}
T.~Lesort, N.~Díaz-Rodríguez, J.-F. Goudou, and D.~Filliat, ``State
  representation learning for control: An overview,'' \emph{Neural Networks},
  vol. 108, pp. 379 -- 392, 2018.

\bibitem{NIPS2018_7512}
D.~Ha and J.~Schmidhuber, ``Recurrent world models facilitate policy
  evolution,'' in \emph{Advances in Neural Information Processing Systems
  31}.\hskip 1em plus 0.5em minus 0.4em\relax Curran Associates, Inc., 2018,
  pp. 2450--2462.

\bibitem{E2C}
M.~Watter, J.~T. Springenberg, J.~Boedecker, and M.~Riedmiller, ``Embed to
  control: A locally linear latent dynamics model for control from raw
  images,'' in \emph{Proceedings of the 28th International Conference on Neural
  Information Processing Systems - Volume 2}, ser. NIPS'15.\hskip 1em plus
  0.5em minus 0.4em\relax MIT Press, 2015, pp. 2746--2754.

\bibitem{hafner2019planet}
D.~Hafner, T.~Lillicrap, I.~Fischer, \emph{et~al.}, ``Learning latent dynamics
  for planning from pixels,'' in \emph{International Conference on Machine
  Learning}, 2019, pp. 2555--2565.

\bibitem{lee2019slac}
A.~X. Lee, A.~Nagabandi, P.~Abbeel, and S.~Levine, ``Stochastic latent
  actor-critic: Deep reinforcement learning with a latent variable model,''
  \emph{arXiv preprint arXiv:1907.00953}, 2019.

\bibitem{Eslami2018NeuralSR}
S.~M.~A. Eslami, D.~J. Rezende, F.~Besse, \emph{et~al.}, ``Neural scene
  representation and rendering,'' \emph{Science}, vol. 360, pp. 1204--1210,
  2018.

\bibitem{NIPS2019_9331}
J.~Tobin, W.~Zaremba, and P.~Abbeel, ``Geometry-aware neural rendering,'' in
  \emph{Advances in Neural Information Processing Systems 32}.\hskip 1em plus
  0.5em minus 0.4em\relax Curran Associates, Inc., 2019, pp. 11\,559--11\,569.

\bibitem{laskin_srinivas2020curl}
M.~Laskin, A.~Srinivas, and P.~Abbeel, ``Curl: Contrastive unsupervised
  representations for reinforcement learning,'' \emph{Proceedings of the 37th
  International Conference on Machine Learning, Vienna, Austria, PMLR 119},
  2020.

\bibitem{vae}
D.~P. Kingma and M.~Welling, ``Auto-encoding variational bayes,'' in \emph{2nd
  International Conference on Learning Representations, {ICLR} 2014, Banff, AB,
  Canada}, 2014.

\bibitem{ku2018defense}
J.~Ku, A.~Harakeh, and S.~L. Waslander, ``In defense of classical image
  processing: Fast depth completion on the cpu,'' in \emph{2018 15th Conference
  on Computer and Robot Vision (CRV)}.\hskip 1em plus 0.5em minus 0.4em\relax
  IEEE, 2018, pp. 16--22.

\bibitem{lstm}
S.~Hochreiter and J.~Schmidhuber, ``Long short-term memory,'' \emph{Neural
  computation}, vol.~9, pp. 1735--1780, 1997.

\bibitem{kalman}
R.~E. Kalman, ``A new approach to linear filtering and prediction problems,''
  \emph{ASME Journal of Basic Engineering}, 1960.

\bibitem{dreaming}
A.~J. Piergiovanni, A.~Wu, and M.~S. Ryoo, ``Learning real-world robot policies
  by dreaming,'' in \emph{2019 {IEEE/RSJ} International Conference on
  Intelligent Robots and Systems, {IROS} 2019, Macau, SAR, China}.\hskip 1em
  plus 0.5em minus 0.4em\relax {IEEE}, 2019, pp. 7680--7687.

\bibitem{seq2seq}
I.~Sutskever, O.~Vinyals, and Q.~V. Le, ``Sequence to sequence learning with
  neural networks,'' in \emph{Advances in Neural Information Processing Systems
  27}.\hskip 1em plus 0.5em minus 0.4em\relax Curran Associates, Inc., 2014,
  pp. 3104--3112.

\bibitem{seq2seq-video}
N.~Srivastava, E.~Mansimov, and R.~Salakhudinov, ``Unsupervised learning of
  video representations using lstms,'' in \emph{Proceedings of the 32nd
  International Conference on Machine Learning}, ser. Proceedings of Machine
  Learning Research, vol.~37.\hskip 1em plus 0.5em minus 0.4em\relax PMLR,
  07--09 Jul 2015, pp. 843--852.

\bibitem{8260889}
C.~D. {Bellicoso}, F.~{Jenelten}, C.~{Gehring}, and M.~{Hutter}, ``Dynamic
  locomotion through online nonlinear motion optimization for quadrupedal
  robots,'' \emph{IEEE Robotics and Automation Letters}, vol.~3, no.~3, pp.
  2261--2268, 2018.

\bibitem{adam}
D.~P. Kingma and J.~Ba, ``Adam: {A} method for stochastic optimization,'' in
  \emph{3rd International Conference on Learning Representations, {ICLR} 2015}.

\bibitem{PPO}
J.~Schulman, F.~Wolski, P.~Dhariwal, A.~Radford, and O.~Klimov, ``Proximal
  policy optimization algorithms,'' \emph{CoRR}, vol. abs/1707.06347, 2017.

\end{thebibliography}

\end{document}